\def\year{2020}\relax
\documentclass[letterpaper]{article} 
\usepackage{aaai20}  
\usepackage{times}  
\usepackage{helvet} 
\usepackage{courier}  
\usepackage[hyphens]{url}  
\usepackage{graphicx} 
\usepackage{graphicx}  
\newcommand{\bv}{\textbf{v}}

\usepackage{todonotes}

\newcommand{\citet}[1]{\citeauthor{#1} \shortcite{#1}}
\newcommand{\citep}{\cite}

\usepackage{booktabs}
\usepackage{amsmath}
\usepackage{breakcites}
\usepackage{latexsym}
\usepackage{amssymb}
\usepackage{array}
\usepackage{bbm}
\usepackage{lipsum}
\usepackage{latexsym}
\usepackage{listings}
\usepackage{enumitem}
\usepackage{todonotes}

\DeclareMathOperator*{\argmax}{arg\,max}

\title{Fine-Grained Entity Typing for Domain Independent Entity Linking}
\author{Yasumasa Onoe \and Greg Durrett \\
Department of Computer Science\\
The University of Texas at Austin\\
{\tt \{yasumasa, gdurrett\}@cs.utexas.edu}
}
\begin{document}
\maketitle

\begin{abstract}
Neural entity linking models are very powerful, but run the risk of overfitting to the domain they are trained in. For this problem, a ``domain'' is characterized not just by genre of text but even by factors as specific as the particular distribution of entities, as neural models tend to overfit by memorizing properties of frequent entities in a dataset. We tackle the problem of building robust entity linking models that generalize effectively and do not rely on labeled entity linking data with a specific entity distribution. Rather than predicting entities directly, our approach models fine-grained entity properties, which can help disambiguate between even closely related entities. We derive a large inventory of types (tens of thousands) from Wikipedia categories, and use hyperlinked mentions in Wikipedia to distantly label data and train an entity typing model. At test time, we classify a mention with this typing model and use soft type predictions to link the mention to the most similar candidate entity. We evaluate our entity linking system on the CoNLL-YAGO dataset \cite{Johannes_Hoffart_11} and show that our approach outperforms prior domain-independent entity linking systems. We also test our approach in a harder setting derived from the WikilinksNED dataset \cite{Yotam_Eshel_17} where all the mention-entity pairs are unseen during test time. Results indicate that our approach generalizes better than a state-of-the-art neural model on the dataset.
\end{abstract}

\section{Introduction}

Historically, systems for entity linking to Wikipedia relied on heuristics such as anchor text distributions \cite{Silviu_Cucerzan_07,David_Milne_08},  tf-idf \cite{Lev_Ratinov_11}, and Wikipedia relatedness of nearby entities \cite{Johannes_Hoffart_11}.  These systems have few parameters, making them relatively flexible across domains. More recent systems have typically been parameter-rich neural network models \cite{Yaming_Sun_15, Ikuya_Yamada_16,Matthew_Francis_Landau_16,Yotam_Eshel_17}. Many of these models are trained and evaluated on data from the same domain such as the CoNLL-YAGO dataset \cite{Johannes_Hoffart_11} or WikilinksNED \cite{Yotam_Eshel_17,David_Mueller_18}, for which the training and test sets share similar distributions of entities. These models partially learn to attain high performance by memorizing the entity distribution of the training set rather than learning how to link entities more generally. As a result, apparently strong systems in one domain may not generalize to other domains without fine-tuning.

In this work, we aim to use feature-rich neural models for entity linking\footnote{Throughout this work, when we say entity linking, we refer to the task of disambiguating a \emph{given} entity mention, not the full detection and disambiguation task which this sometimes refers to.} in a way that can effectively generalize across domains. We do this by framing the entity linking problem as a problem of prediction of very fine-grained entity types. Ambiguous entity references (e.g., different locations with the same name, the same movie released in different years) often differ in critical properties that can be inferred from context, but which neural bag-of-words and similar methods may not effectively tease out. We use an inventory of tens of thousands of types to learn such highly specific properties. This represents a much larger-scale tagset than past work using entity typing for entity linking, which has usually used hundreds of types \cite{Nitish_Gupta_17,Shikhar_Murty_18,Jonathan_Raiman_18}. Critically, type prediction is the only learned component of our model: our final entity prediction uses a very simple heuristic based on summing posterior type probabilities. 


\begin{figure*}[!t]
    \centering
    \includegraphics[width=1.0\linewidth]{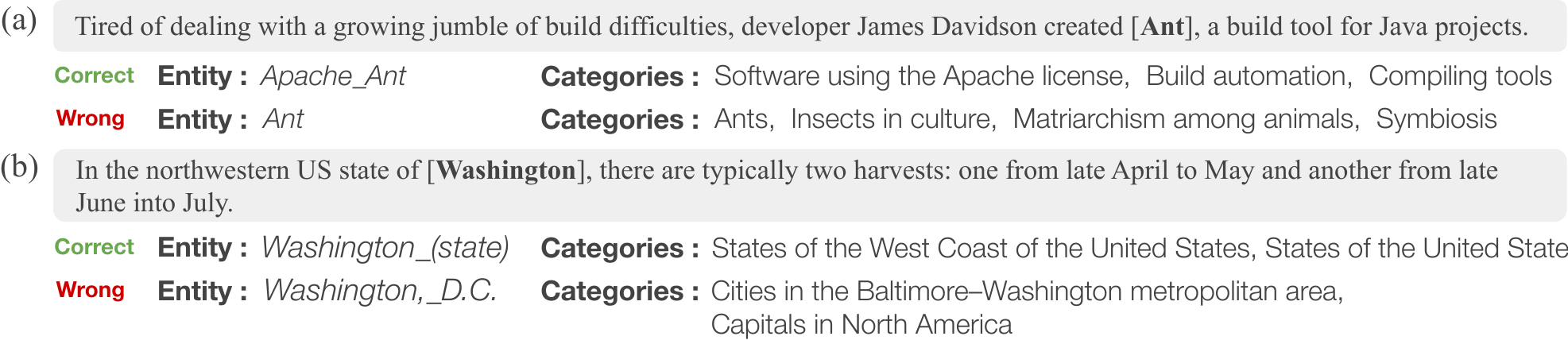}
    \caption{Examples selected from the WikilinksNED development set \protect \cite{Yotam_Eshel_17}. The mention (in bold) resolves to the topmost entity in each case. These correct entities can be distinguished by their fine-grained Wikipedia categories.}
    \label{fig:motivation}
\end{figure*}


To train our typing model, we collect data from Wikipedia targeting a range of types in a domain of interest. This type set is lightly specialized to the target domain, but importantly, the set is determined on the basis of purely unlabeled data in the domain (lists of candidates for the identified mentions). Moreover, because we use such a large type inventory, our model captures a wide range of types and can handle entity linking in both narrow settings such as CoNLL-YAGO and broader domain settings such as WikilinksNED. Our typing model itself is adapted from past work on ultra-fine grained entity typing in a different setting \cite{Eunsol_Choi_18,Onoe_Durrett_19}. As a high-capacity neural network model, this model can train on millions of examples and effectively predict even rare types.

Our contributions are as follows: (1) Formulating entity linking as purely an entity typing problem. (2) Constructing a distantly-supervised typing dataset based on Wikipedia categories and training an ultra-fine entity typing model on it. (3) Showing through evaluation on two domains that our model is more effective than a range of other approaches trained from out-of-domain data, including Wikipedia data specialized to that particular domain.

\section{Motivation and Setup}

Figure~\ref{fig:motivation} shows two examples from the WikilinksNED development set \cite{Yotam_Eshel_17} which motivate our problem setup. In the example (a), the most frequent Wikipedia entity given the mention ``Ant'' is the insect ant. In the Wikipedia dump, source anchors ``Ant'' point to the Wikipedia article about the insect {\it Ant}\footnote{We use italics to denote \emph{Wikipedia titles} and true type to represent \texttt{Wikipedia categories}.} $96\%$ of the time and points to {\it Apache\_Ant} $0.8\%$ of the time. Since {\it Apache\_Ant} is very rare in the training data, models trained on broad domain data will often prefer {\it Ant} in many contexts.

Predicting categories here is much less problematic. Our category processing (described in the Training Data for Typing section) assigns this mention several categories including {\tt Software}. Predicting {\tt Software} in this context is relatively easy for a typing model given the indicative words in the context. Knowing this type is enough to disambiguate between these two entities independent of other clues, and notably, it is not skewed by the relative rarity of the {\it Apache\_Ant} title. The category information is shared across many entities, so we can expect that predicting the category information would be much more efficient than learning rare entities directly. 

The example (b) adds another challenge since ``Washington'' can correspond to many different people or locations, some of which can occur in relatively similar contexts. Even a coarse-grained type inventory will distinguish between these mentions. However, more specific category information is needed to distinguish between {\it Washington\_(state)} and {\it Washington,\_D.C.} In this case, {\tt States of the West Coast of the United States} would disambiguate between these, and context clues like ``northwestern'' can help identify this. This category is extremely fine-grained and we cannot assume an ability to predict it reliably; we discuss in the Training Data for Typing section how to get around this limitation by splitting categories into parts.

Finally, we note that a global linking system \cite{Johannes_Hoffart_11} can sometimes exploit relevant context information from related entities like Java (programming language). In this model, we focus on a purely local approach for simplicity and see how far this can go; this approach is also the most general for datasets like WikilinksNED where other reliable mentions may not be present close by.

\paragraph{Setup} We focus on the entity linking (EL) task of selecting the appropriate Wikipedia entities for the mentions in context. We use $m$ to denote a mention of an entity, $s$ to denote a context sentence, $e$ to denote a Wikipedia entity associated with the mention $m$, and $C$ to denote a set of candidate entities. We also assume that we have access to a set of Wikipedia categories $T$ corresponding to the Wikipedia entity $e$.

Suppose we have an entity linking dataset $\mathcal{D}_{\text{EL}} = \{(m, s, e, C)^{(1)}, \dots, (m, s, e, C)^{(k)}\}$. In the standard entity linking setting, we train a model using the training set of $\mathcal{D}_{\text{EL}}$ and evaluate on the development/test sets of $\mathcal{D}_{\text{EL}}$. In our approach, we also have an entity typing dataset collected from hyperlinks in English Wikipedia $\mathcal{D}_{\text{Wiki}} = \{(m, s, T)^{(1)}, \dots, (m, s, T)^{(l)}\}$. Since $\mathcal{D}_{\text{Wiki}}$ is derived from Wikipedia itself, this data contains a large number of common Wikipedia entities. This enables us to train a general entity typing model that maps the mention $m$ and its context $s$ to a set $T$ of Wikipedia categories: $\Phi: (m,s) \rightarrow T$. Then, we use a scoring function $\Omega$ to make entity linking predictions based on the candidate set: $e = \Omega(\Phi(m, s), C)$. We evaluate our approach on the development/test sets of the existing entity linking data $\mathcal{D}_{\text{EL}}$.
During training, we never assume access to labeled entity data $\mathcal{D}_{\text{EL}}$. Furthermore, by optimizing to predict Wikipedia categories $T$ instead of an entity $e$, we can achieve a higher level of generalization across entities rather than simply memorizing our Wikipedia training data.

\section{Model}\label{model}

Our model consists of two parts: a learned entity typing model and a heuristic (untrained) entity link predictor that depends only on the types.


\begin{figure*}[!t]
\centering
    \centering
    \includegraphics[width=1.0\linewidth]{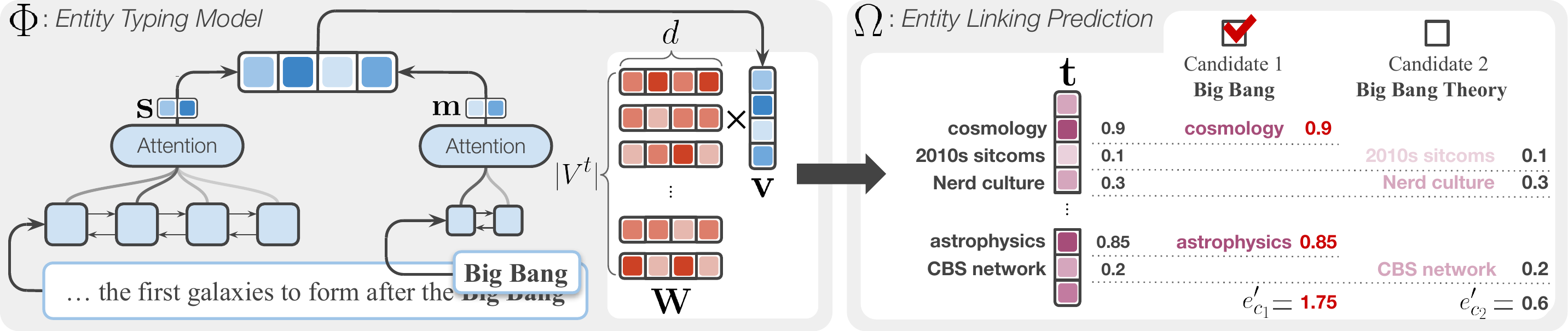}
    \caption{Entity typing for entity linking (ET4EL) model. Given a mention $m$ and a sentence $s$, the entity typing model $\Phi$ computes a binary probability for membership in each type. Then the entity linking predictor $\Omega$ makes the final prediction based on summed type posteriors: the model chooses Big Bang over Big Bang Theory based on these scores (1.75 vs 0.6).}
    \label{fig:model}
\end{figure*}


\subsection{Entity Typing Model}

Figure~\ref{fig:model} summarizes the model architecture of the entity typing model $\Phi$. We use an attention-based model \cite{Onoe_Durrett_19} designed for the fine-grained entity typing tasks \cite{Dan_Gillick_14, Eunsol_Choi_18}. This model takes a mention span in a sentence context, uses span attention over vector representations of the mention span and its context, then aggregates that information to predict types. We follow \citet{Onoe_Durrett_19} for our entity typing model design and hyperparameter choices.

\paragraph{Encoder} 
The mention $m$ and the sentence $s$ are converted into sequences of contextualized word vectors $s'$ and $m'$ using ELMo \cite{ELMO_18}. The sentence vectors $s'$ are concatenated with the location embedding $\ell$  and fed into a bi-LSTM encoder followed by a span attention layer \cite{Kenton_Lee17,Eunsol_Choi_18}: $\mathbf{s} = \text{Attention}(\text{bi-LSTM}([s'; \ell]))$, where $\mathbf{s}$ is the final representation of the sentence $s$. The mention vectors $m'$ are fed into another bi-LSTM and summed by a span attention layer to obtain the word-level mention representation: $\mathbf{m}^{\text{word}} = \text{Attention}(\text{bi-LSTM}(m'))$. We also use a 1-D convolution over the characters of the mention to generate a character-level mention representation  $\mathbf{m}^{\text{char}}$. The final representation of the mention and the sentence is a concatenation of the three vectors: $\bv = \left[\mathbf{s}; \mathbf{m}^{\text{word}}; \mathbf{m}^{\text{char}} \right] \in \mathbb{R}^d$. Unlike \citet{Onoe_Durrett_19}, we do not include the contextualized word vectors of the mention headword.\footnote{Compared to other models we considered, such as BERT \cite{devlin19}, this approach was more stable and more scalable to use large amounts of Wikipedia data.} 

\paragraph{Decoder} We use $|V^t|$ to denote the size of the category vocabulary. Following previous work \cite{Eunsol_Choi_18,Onoe_Durrett_19}, we assume independence between the categories; thus, this boils down to a binary classification problem for each of the categories. The decoder is a single linear layer parameterized with $\mathbf{W} \in \mathbb{R}^{|V^t| \times d}$. The probabilities for all categories in the vocabulary are obtained by $\mathbf{t} = \sigma(\mathbf{W} \mathbf{v})$, where $\sigma(\cdot)$ is an element-wise sigmoid operation. The probability vector $\mathbf{t}$ is the final output from the entity typing model $\Phi$. 

\paragraph{Learning} The entity typing model $\Phi$ is learned on the training examples consisting of $(m,s,T)$ triples. The loss is a sum of binary cross-entropy losses over all categories over all examples. That is, the typing problem is viewed as independent classification for each category, with the mention-context encoder shared across categories. Formally, we optimize a multi-label binary cross entropy objective:
\begin{equation}
\label{loss1}
\begin{aligned}
\mathcal{L} = -\sum_{i} y_i \cdot \log t_i + (1 - y_i) \cdot \log (1 - t_i),\\
\end{aligned}
\end{equation}
where $i$ are indices over categories, $t_i$ is a score of the $i$th category, and $y_i$ takes the value 1 if the $i$th category applies to the current mention.

\subsection{Entity Linking Prediction}

Once the entity typing model $\Phi$ is trained, we use the model output $\mathbf{t}$ to make entity linking predictions. Assume we have an example from the test set of an entity linking dataset: $x = (m, s, C)$, where $C$ is a set of the candidate entities. We perform the forward computation of $\Phi$ and obtain the probability vector $\mathbf{t} = \Phi(m, s)$. Then, we score all the candidates in $C$ using a scoring function $\Omega$. Our choice of $\Omega$ is defined as the sum of probabilities for each type exhibited by the selected entity:
\begin{equation}
\label{omega}
\begin{aligned}
e'_c =& \sum_{i} t_i \cdot \mathbbm{1}_{T_c}\left(V^t_i\right)\\
e =& \argmax_{e} \left(e'_1, \dots, e'_{|C|} \right),
\end{aligned}
\end{equation}
where $e'_c$ is a score for a candidate entity $c$, $\mathbbm{1}_{T_c}\left(\cdot\right)$ is an indicator function that is activated when $i$th category in the vocabulary $V^t_i$ is in the set of categories of the candidate entity $c$, and $e$ is a predicted entity. 

We observed that simply summing up the scores performed better than other options such as averaging or computing a log odds ratio. Intuitively, summing benefits candidates with many categories, which biases the model towards more frequent entities in a beneficial way. It also rewards models with many correlated types, which is problematic, but approaches we tried that handled type correlation in a more principled way did not perform better.

There are certain types of entities in Wikipedia whose categories do not mesh well with our prediction task. For example, the page \emph{Employment} about the general concept only has the category \texttt{Employment}, making resolution of this concept challenging. In these cases, we back off to a mention-entity prior (see in the Preprocessing Evaluation Data section). We call our combined system ET4EL (entity typing for entity linking).\footnote{The code for experiments is available at \url{https://github.com/yasumasaonoe/ET4EL}}

\begin{table*}[!t]
	\centering
	\small
	\begin{tabular}{l c c c}
		\toprule
		\multicolumn{1}{c}{Model}   & \multicolumn{1}{c}{Input} & \multicolumn{1}{c}{Training Data} & \multicolumn{1}{c}{Supervision} \\
		\midrule
		{\sc ET4EL} (this work) & mention, context & Wiki & mention-categories \\
		\citet{Nitish_Gupta_17} CDTE & document &  Wiki  & mention-entity \\
		\citet{Lazic_Nevena_15} Plato, sup & document & Wiki & mention-entity\\
		\citet{Lazic_Nevena_15} Plato, semi-sup & document & Wiki + 50M Web pages  & mention-entity\\
		\citet{Phong_Le_19}  & document & Wiki + 30k RCV1 docs & mention-entity\\
		Standard EL Systems (local) & mention, context & domain-specific training set  & mention-entity\\
		Standard EL Systems (global) & document &  domain-specific training set  & mention-entity\\
		\bottomrule 
	\end{tabular}
	\caption{Assumptions and resources for different entity linking systems. Our model only requires supervision from Wikipedia and trains using typing supervision (from categories) only.}
	\label{tab:systems}
\end{table*}

\section{Training Data for Typing}\label{data}
To train the ET4EL model to cover a wide range of entities, we need access to a large set of entities labeled with types. We derive this data directly from Wikipedia: each hyperlinked occurrence of an entity on Wikipedia can be treated as a distantly supervised \cite{Mark_Craven_99,Mike_Mintz_09} example from the standpoint of entity typing. The distant types for that mention are derived from the Wikipedia categories associated with the linked entity.

\paragraph{Annotation} First, we collect all sentences that contain hyperlinks, internal links pointing to other English Wikipedia articles, from all articles on English Wikipedia. Our data is taken from the March 2019 English Wiki dump. Given a sentence with a hyperlink, we use the hyperlink as a mention $m$, the whole sentence as a context sentence $s$, the destination of the hyperlink as an entity $e$, and the Wiki categories that are associated with $e$ as a set of fine-grained types $T$. One sentence could have multiple hyperlinks. In this case, we create a tuple $(m, s, e, T)$ for each of the hyperlinks. This process results 88M examples. Importantly, our training examples for typing are tuples of $(m, s, T)$ since the ET4EL model is optimized towards the gold Wiki categories $T$ and do not rely on the gold entity $e$. 

\paragraph{Category Set} The original Wikipedia categories are mostly fine-grained and lack general categories. For example, the Wiki entity {\it New\_York\_ City} has fine-grained categories such as {\tt Cities in New York (state)} and {\tt Populated places established in 1624}, but there are no general categories (e.g. {\tt Cities}) that potentially useful to distinguish between two obviously different entities (e.g. location vs person). We expand the original categories if they contain prepositions.\footnote{We use `in', `from', `for', `of', `by', `for', `involving.'} We split the original category at the location of the first-occurring preposition. We chunk the left side into words and add them to the category set. We add the right side, a prepositional phrase, to the category set without modification; retaining the preposition helps keep the relation information. We also retain the original category. For the two original categories above, the new categories {\tt Cities, in New York (state), Populated, places, established, in 1624} would be added to the category set.\footnote{Other splits are possible, e.g. extracting \emph{20th century} from \emph{20th century philosophers}. However, these are more difficult to reliably identify.} 

Past work has looked at deriving similar category sets over Wikipedia \cite{Nastase08}. Further improvements to our category set are possible, but we found the simple of rules we defined to be sufficient for our purposes.

\paragraph{Training the Typing Model} Since the total number of Wikipedia categories is very large (over 1 million), we train on a subset of the categories for efficiency. For a given test set, we only need access to categories that might possibly occur. We therefore restrict the categories to the most common $n$ categories occurring with candidates in that dataset; note that this does not assume the existence of labeled target-domain data, only unlabeled.

To create the training set, we randomly select 6M examples that have at least one Wikipedia category from the restricted category vocabulary. We select 10k examples for the development set using the same procedure. The encoder may specialize somewhat to these types, but as we show later, it can handle large type sets and recognize diverse entity types (see the Results and Discussion section).

\section{Experiments}\label{experiments}

We evaluate our approach on the development/test sets of the CoNLL-YAGO \cite{Johannes_Hoffart_11} dataset, which is a widely used entity linking benchmark. The CoNLL data consists of news documents and covers relatively narrow domains. Additionally, we test our model in a much harder setting where the mention-entity pairs are unseen during test time. We create the training, development, and test sets from the WikilinksNED dataset \cite{Yotam_Eshel_17}, which contains a diverse set of ambiguous entities spanning more domains than the CoNLL data. We call this dataset Unseen-Mentions. The domain-specific training set is only used for the baseline models. The ET4EL model is still trained on the Wikipedia data. Unlike the CoNLL data, the examples in the Unseen-Mentions dataset are essentially single-sentence, meaning that resolution has to happen with limited context.

\subsection{Preprocessing Evaluation Data}\label{prep_eval} 

\paragraph{Candidate Selection} For the CoNLL data, we use the publicly available candidate list, PPRforNED \cite{Maria_Pershina_15} that gives 99\% gold recall on the testa (development) and the testb (test) sets.\footnote{Other domain independent entity linking systems employ different resources to generate candidates, e.g., \citet{Nitish_Gupta_17} use CrossWikis \cite{spitkovsky12} and restrict to 30 candidates per mention (for 95\% gold recall). \citet{Lazic_Nevena_15} use the Wikilinks corpus, Wikipedia articles, and Freebase (for 92\% gold recall). Because all these systems make slightly different precision-recall tradeoffs in their candidate selection, direct comparison is difficult, but we believe the results are still reflective of the overall quality of the systems.}

For the Unseen-Mentions data, we use a mention-entity prior $\hat p (e | m)$ to select candidate entities \cite{ganea_hofmann_17}. We compute $\hat p (e | m)$ using the count statistics from the Wiki dump. We rank the candidate entities based on $\hat p (e | m)$ and clip low frequency entities with a threshold $0.05$. On average, this produces around 30 candidates per example and gives 88\% gold recall on the development and test sets.

\paragraph{Category Vocabulary} To reduce more than 1 million total Wikipedia categories to a more tractable number for a given dataset, we use count statistics from the \emph{candidates} of the training examples in that data set. Note that this process does not use the training labels at all; the data may as well be unlabeled. For each category, we count the number of associated unique mentions. We rank all the categories by the counts and select the top 60k categories as our vocabulary.

\subsection{Baselines}

\begin{description}[align=left, font=\sc, topsep=2pt, labelsep=\fontdimen2\font, leftmargin=0pt]

\item [Most Frequent Entity] \hspace{2pt}  Given a mention $m$, we choose an entity with the highest  mention-entity prior $\hat p (e | m)$. We compute $\hat p (e | m)$ using the count statistics from the March 2019 Wiki dump. 
\item [Cosine Similarity] \hspace{2pt}  This baseline selects an entity with the highest cosine similarity between the context and entity vectors using the pretrained {\tt word2vecf} \cite{Omer_Levy_14}. The context vector is obtained by mean pooling over the input word vectors. Note that this similarity is computed using distributed representations while traditional cosine similarity is based on word counts \cite{Johannes_Hoffart_11}.
\item [GRU-Attn] \hspace{2pt} Our implementation of the attention-based model introduced in \citet{David_Mueller_18}. This model achieves state-of-the-art performance on the WikilinksNED dataset in the standard supervised setting. See \citet{David_Mueller_18} for more details.
\item [CBoW+word2vec] \hspace{2pt} This simpler baseline model\footnote{This model shows comparable performance to {\sc GRU-Attn}, achieving $76.0$ accuracy on the original test set of the WikilinksNED data, comparable to the performance of $75.8$ reported in \citet{David_Mueller_18}.} uses the pretrained embeddings and a simple bag-of-words mention-context encoder. That is, the encoder is unordered bag-of-words representations of the mention, the left context, and the right context. For each of three, the words are embedded and combined using mean pooling to give context representations. Similar to \citet{Yotam_Eshel_17}, we use {\tt word2vecf} to initialize entity embeddings. We compare the context representations and the entity representation by following \citet{David_Mueller_18}. The final representation is fed into a two-layer multilayer perceptron (MLP) with ReLU activations, batch normalization, and dropout.
\end{description}

\paragraph{Training Data for CoNLL Baselines} To train our baselines in a comparable fashion, we create training examples $(m, s, e)$ from the Wikipedia data $\mathcal{D}_{\text{Wiki}}$ to use for our learning-based baselines ({\sc GRU-Attn} and {\sc CBoW+word2vec}). We use the same mention-entity prior $\hat p (e | m)$, explained in the previous section, to select candidates for each training example.

We consider two variants of this training data. First, we train these baselines on a training set sampled uniformly from all of Wikipedia. Second, we give these baselines a more favorable transductive setting where the training entities from Wikipedia are restricted to only include entities that are candidates in the domain-specific training data. The CoNLL training set contains 2k unique entities. We collect 1.4M training examples from $\mathcal{D}_{\text{Wiki}}$ that cover these 2k CoNLL entities; this training set should specialize these models to CoNLL fairly substantially, though they maintain our setting of not considering the training labels. 

\paragraph{Training Data for Unseen-Mentions Baselines}

To ensure that all mentions in the development and test sets do not appear in the training set, we split the WikilinksNED training set into train, development, and test sets by unique mentions (15.5k for train, 1k for dev, and 1k for test). This results 2.2M, 10k, and 10k examples\footnote{Development and test are subsampled from their ``raw'' sizes of 130k token-level examples.}  respectively. Our learning-based baselines ({\sc GRU-Attn} and {\sc CBoW+word2vec}) are trained on the 2.2M training examples, which do not share any entities with the dev or test sets.

We also train the learning-based baselines on the Wikipedia data described in the Training Data for Typing section. Similar to the Unseen-Mentions data, we use a mention-entity prior $\hat p (e | m)$ to select candidate entities. We obtain 2.5M training examples that have at least 2 candidate entities.

\paragraph{Comparison with other systems} Table~\ref{tab:systems} compares assumptions and resources for different systems. The ET4EL model is a so-called local entity linking model, as it only uses a single mention and context, rather than a global model which does collective inference over the whole document \cite{Lev_Ratinov_11,Johannes_Hoffart_11}. However, this allows us to easily support entity linking with little context, as is the case for WikilinksNED.

The chief difference from other models is that the ET4EL model is trained on data derived from Wikipedia, where the supervision comes from categories attached to entity mentions. Moreover, we \emph{only} use Wikipedia as a source of training data; some other work like \citet{Phong_Le_19} uses unlabeled data from the same domain as the CoNLL-YAGO test set.

\begin{table}[!t]
	\centering
	\small
	\begin{tabular}{lccc}
		\toprule
		\multicolumn{1}{c}{Model} & {Dev} & {Test} \\
		\midrule
		{\sc Most Frequent Entity}& 57.7 & 57.3 \\
		{\sc Cosine Similarity} & 45.5 & 42.8\\ 
		{\sc GRU+Attn} \cite{David_Mueller_18} & 67.5 & 63.2 \\
		{\sc GRU+Attn} (Transduction) & 82.0 & 75.3 \\
		{\sc CBoW+word2vec} & 70.1 & 67.3 \\
		{\sc CBoW+word2vec} (Transduction) & 84.6 & 77.5 \\
		{\sc ET4EL} (this work) & \textbf{88.1} & \textbf{85.9} \\
		\midrule
		\citet{Nitish_Gupta_17} CDTE & 84.9 & 82.9 \\
	    \citet{Lazic_Nevena_15} Plato, sup & - & 79.7 \\
		\citet{Lazic_Nevena_15} Plato, semi-sup\textsuperscript{\textdagger}& - & 86.4 \\ 
		\citet{Phong_Le_19}\textsuperscript{$\ddagger$}  & - & 89.7 \\ 
		\bottomrule 
	\end{tabular}
	\caption{Accuracy on the CoNLL development set (testa) and the CoNLL test set (testb). \textdagger: trained with additional large unlabeled data. $\ddagger$: uses in-domain unlabeled data (RCV1). Our model outperforms the baselines and models using similar data from prior work.}
	\label{tab:CoNLL-main}
\end{table}

\begin{table}[!t]
	\centering
	\small
	\begin{tabular}{lcc}
		\toprule
		\multicolumn{1}{c}{Amount of Context} & {Dev} \\
		\midrule
		Sentence  & 83.8 \\
		Sentence + left \& right 50 tokens  & 84.5 \\
		Sentence + 1st doc sentence  & \textbf{88.1} \\
		\bottomrule 
	\end{tabular}
	\caption{Accuracy on the CoNLL development set (testa) with different amounts of context fed to our model. Adding the first sentence of the document gives the best performance because this is often indicative of topic in this dataset (e.g., what sport is being discussed). }
	\label{tab:CoNLL-inp}
\end{table}

\begin{table}[!t]
	\centering
	\small
	\begin{tabular}{lcccc}
		\toprule
		Category Size & 1k & 5k & 30k & 60k \\
		\midrule
		Dev & 85.1 & 85.6 & 87.1 & \textbf{88.1}\\
		\bottomrule 
	\end{tabular}
	\caption{Accuracy on the CoNLL development set (testa) with different numbers of categories. }
	\label{tab:CoNLL-vocabsize}
\end{table}
\section{Results and Discussion}

\subsection{CoNLL-YAGO}

Table~\ref{tab:CoNLL-main} shows accuracy of our model and baselines. Our model outperforms all baselines by a substantial margin. The {\sc Most Frequent Entity} baseline performs poorly on both development and test set. Interestingly, the simpler {\sc CBoW+word2vec} model is the strongest baseline here, outperforming the {\sc GRU+Attn} model in both general and transductive settings. Our model achieves the strongest performance on both the dev and test data. Interestingly, our model also has a much smaller drop from dev to test, only losing 2 points, compared to the transductive models, which drop by 7 points. The CoNLL testb set is slightly ``out-of-domain'' for the training set with respect to the time period it was drawn from, indicating that our method may have better generalization than more conventional neural models in the transductive setting.

We also list the state-of-the-art domain independent entity linking systems. Our model outperforms the full CDTE model of \citet{Nitish_Gupta_17}, as well as Plato in the supervised setting \cite{Lazic_Nevena_15}, which is the same setting as ours. Our model is competitive with Plato in the semi-supervised setting, which additionally uses 50 million documents as unlabeled data. \citet{Phong_Le_19}'s setting is quite different from ours that their model is a global model (requires document input) and trained on Wikipedia and 30k newswire documents from the Reuters RCV1 corpus \cite{Lewis_04}. Their model is potentially trained on domain-specific data since the CoNLL-YAGO dataset is derived from the RCV1 corpus.

\begin{table}[!t]
	\centering
	\small
	\begin{tabular}{lccc}
		\toprule
		\multicolumn{1}{c}{Model}  & {Training} 
		 & {Test} \\
		\midrule
		{\sc Most Frequent Entity} & Wiki & 54.1 \\
		{\sc Cosine Similarity} & Wiki & 21.7\\
		{\sc GRU+Attn} \cite{David_Mueller_18} & in-domain & 41.2 \\
		{\sc GRU+Attn} & Wiki & 43.4 \\
		{\sc CBoW + word2vec}\textsuperscript{\textdagger} & in-domain & 43.0 \\
		{\sc CBoW + word2vec} & Wiki & 38.0 \\
		{\sc ET4EL} (this work) & Wiki & \textbf{62.2} \\
		\bottomrule 
	\end{tabular}
	\caption{Accuracy on the Unseen-Mentions test set. Our model substantially outperforms neural entity linking models in this setting.}
	\label{tab:wikilinks}
\end{table}

\paragraph{How much context information should we add?}  On the CoNLL dataset, sentences containing entity mentions are often quite short, but are embedded in a larger document. We investigate the most effective amount of context information to add to our typing model. Table~\ref{tab:CoNLL-inp} compares accuracy for the different amount of context. We test the context sentence only, the left and right 50 tokens of the mention span, and the first sentence of the document. Adding the left and right 50 tokens of the mention span improves the accuracy over the context sentence only. Adding the first sentence of the document improves the accuracy over the context sentence only (no additional context) by 4 points.\footnote{Our baselines use this setting as well since we found it to work the best.} Since the documents are news articles, the first sentence usually has meaningful information about the topics. This is especially useful when the document is a list of sports results, and a sentence does not have rich context. For example, one sentence is ``Michael Johnson ( U.S. ) 20.02'', which is highly uninformative, but the first sentence of the document is ``ATHLETICS - BERLIN GRAND PRIX RESULTS.'' Our model correctly predicts {\it Michael\_Johnson\_(sprinter)} after giving more context information about the sport.

\begin{table*}[!t]
	\centering
	\small
	\begin{tabular}{l c c c c c c c c c c c c c c c}
		\toprule
		\multicolumn{1}{c}{} & \multicolumn{3}{c}{Total} & \multicolumn{3}{c}{1-100} & \multicolumn{3}{c}{101-500} & \multicolumn{3}{c}{501-10000} & \multicolumn{3}{c}{10001+} \\
	    \cmidrule(r){2-4}  \cmidrule(r){5-7} \cmidrule(r){8-10} \cmidrule(r){11-13} \cmidrule(r){14-16}
		\multicolumn{1}{c}{Model}
		 & P & R & F1  & P & R & F1 & P & R & F1 & P & R & F1 & P & R & F1\\
		\midrule
		{\sc ET4EL} (this work)   & 76.2 & 46.1 & 57.5  & 79.7 & 61.3 & 69.3 & 79.0 & 39.9 & 53.0 & 76.2 & 40.1 & 52.5 & 76.5 & 37.0 & 49.9\\
		\bottomrule 
	\end{tabular}
	\caption{Macro-averaged P/R/F1. Entity typing performance on the categories grouped by frequency. (1-100) is the most frequent group, and (10001+) is the least frequent group.}
	\label{tab:typing}
\end{table*}

\paragraph{Does the category vocabulary size matter?} We show the performance on the development set with different numbers of categories.  As we can see in Table~\ref{tab:CoNLL-vocabsize}, the development accuracy monotonically increases as the category size goes up. Even the 1k most frequent category set can achieve reasonable performance, $85\%$ accuracy. However, the model is able to make use of even very fine-grained categories to make correct predictions.

\subsection{Unseen-Mentions}\label{wikilinks-results}
Table~\ref{tab:wikilinks} compares accuracy of our model and baselines on this dataset. Our model achieves the best performance in this setting, better than all baselines. Notably, the {\sc GRU+Attn} model, which achieves state-of-the-art performance on WikilinksNED, performs poorly, underperforming the {\sc Most Frequent Entity} baseline.  The simpler {\sc CBoW+word2vec} model with the frozen entity embeddings shows slightly better performance than the {\sc GRU+Attn} model, implying that the model suffers from overfitting to the training data. The poor performance of these two models trained on the domain-specific data suggests that dealing with unseen mention-entity pairs is challenging even for these vector-based approaches trained with similar domain data, indicating that entity generalization is a major factor in entity linking performance. The {\sc GRU+Attn} model trained on the Wikipedia data also performs poorly.
 
The baseline models trained on the domain-specific data even make mistakes in easy cases such as disambiguating between {\tt PERSON} and {\tt LOCATION} entities. For example, a mention spans is {\bf [Kobe]}, and an associated entity could be {\it Kobe\_Bryant}, a former basketball player, or {\it Kobe}, a city in Japan. Those baseline models guess {\it Kobe\_Bryant} correctly but get confused with {\it Kobe}. Our model predict both entities correctly; the context is usually indicative enough.

\subsection{Typing Analysis}

In the Training Data for Typing section, we described how we added more general types to the category set. We compare the original Wikipedia category set and the expanded category set on the CoNLL development set. Using 30k categories in both settings, the original set and expanded set achieve accuracies of 84.4 and 87.1 respectively, showing that our refined type set helps substantially.

Table~\ref{tab:typing} shows the typing performance on the 60k categories grouped by frequency. The first group (1-100) consists of the 100 most frequent categories. The fourth group (10001+) is formed with the least frequent categories. Precision is relatively high for all groups. The first group (1-100) achieves the highest precision, recall, and F1, possibly leveraging the rich training examples. Recall drastically decreases between the first group and the subsequent groups, which suggests the model has difficulty accounting for the imbalanced nature of the classification of rare tags. 

We further look at the performance of selected individual categories. We observe that having rich training examples, in general, leads the high performance. For example, {\tt births} occurs with more than 2k unique mentions in the training set and achieves P:99/R:89/F1:93.7. However, {\tt history} has more than 900 unique mentions in the training set but only achieves P:76.9/R:6.1/F1:11.4. This might be related to the purity of mentions (and entities). Most of the mentions for {\tt births} are person entities, and this category is consistently applied. On the other hand, {\tt history} may not denote a well-defined semantic type.

\section{Related Work}

Wikipedia categories have been used to construct ontologies \cite{Suchanek07} and predict general concepts \cite{Syed08}. The internal link information in Wikipedia as supervision has also been studied extensively in the field of entity linking and named entity disambiguation in the past decade \cite{Razvan_Bunescu_06, Rada_Mihalcea_07, Joel_Nothman_08, Paul_McNamee_09}. Another approach utilizes manually annotated domain-specific data, using either neural techniques \citep{Zhengyan_He_13,Yaming_Sun_15,Matthew_Francis_Landau_16} or various joint models \cite{durrett_klein_14,Thien_Huu_Nguyen_16}. Learning pretrained entity representations from knowledge bases has also been studied for entity linking \cite{Zhiting_Hu_15,Ikuya_Yamada_16,Ikuya_Yamada_17,Yotam_Eshel_17}. Many of these approaches are orthogonal to ours and could be combined in a real system.

\section{Conclusion}
In this paper, we presented an entity typing approach that addresses the issue of overfitting to the entity distribution of a specific domain. Our approach does not rely on labeled entity linking data in the target domain and models fine-grained entity properties. With the domain independent setting, our approach achieves strong results on the CoNLL dataset. In a harder setting of unknown entities derived from the WikilinksNED dataset, our approach generalizes better than a state-of-the-art model on the dataset.

\section*{Acknowledgments}

This work was partially supported by NSF Grant IIS-1814522, NSF Grant SHF-1762299, a Bloomberg Data Science Grant, and an equipment grant from NVIDIA. The authors acknowledge the Texas Advanced Computing Center (TACC) at The University of Texas at Austin for providing HPC resources used to conduct this research. Results presented in this paper were obtained using the Chameleon testbed supported by the National Science Foundation. Thanks as well to the anonymous reviewers for their thoughtful comments, members of the UT TAUR lab for helpful discussion, Pengxiang Cheng and Jiacheng Xu for constructive suggestions, and Nitish Gupta for providing the details of experiments.

\fontsize{9.0pt}{10.pt} \selectfont
\bibliographystyle{aaai}
\bibliography{aaai}

\end{document}